\documentclass[10pt,journal,compsoc]{IEEEtran}
\usepackage[switch]{lineno}
\usepackage{graphicx}
\usepackage{amsmath}
\usepackage{amssymb}
\usepackage{subcaption}
\usepackage{makecell}
\usepackage{booktabs}
\usepackage{url}
\usepackage{color, colortbl}
\usepackage[dvipsnames]{xcolor}
\usepackage{multirow}
\usepackage{tabularray}
\usepackage{array}

\usepackage{tikz}
\newcommand{\tikzxmark}{%
\tikz[scale=0.23] {
    \draw[line width=0.7,line cap=round] (0,0) to [bend left=6] (1,1);
    \draw[line width=0.7,line cap=round] (0.2,0.95) to [bend right=3] (0.8,0.05);
}}
\newcommand{\tikzcmark}{%
\tikz[scale=0.23] {
    \draw[line width=0.7,line cap=round] (0.25,0) to [bend left=10] (1,1);
    \draw[line width=0.8,line cap=round] (0,0.35) to [bend right=1] (0.23,0);
}}

\newcommand{\PreserveBackslash}[1]{\let\temp=\\#1\let\\=\temp}
\newcolumntype{C}[1]{>{\PreserveBackslash\centering}p{#1}}
\newcolumntype{R}[1]{>{\PreserveBackslash\raggedleft}p{#1}}
\newcolumntype{L}[1]{>{\PreserveBackslash\raggedright}p{#1}}

\def\ie{\emph{i.e.}}
\def\eg{\emph{e.g.}}

\def\etal{\emph{et al.~}}

\definecolor{Gray}{gray}{0.9}
\definecolor{LightCyan}{rgb}{0.88,1,1}

\newcolumntype{a}{>{\columncolor{LightCyan}}c}

\newcommand{\edit}[1]{{\color{Black}{#1}}}

\usepackage{graphicx}
\usepackage{etoolbox}
\newcommand{\insertfig}{
\includegraphics[width=\textwidth]{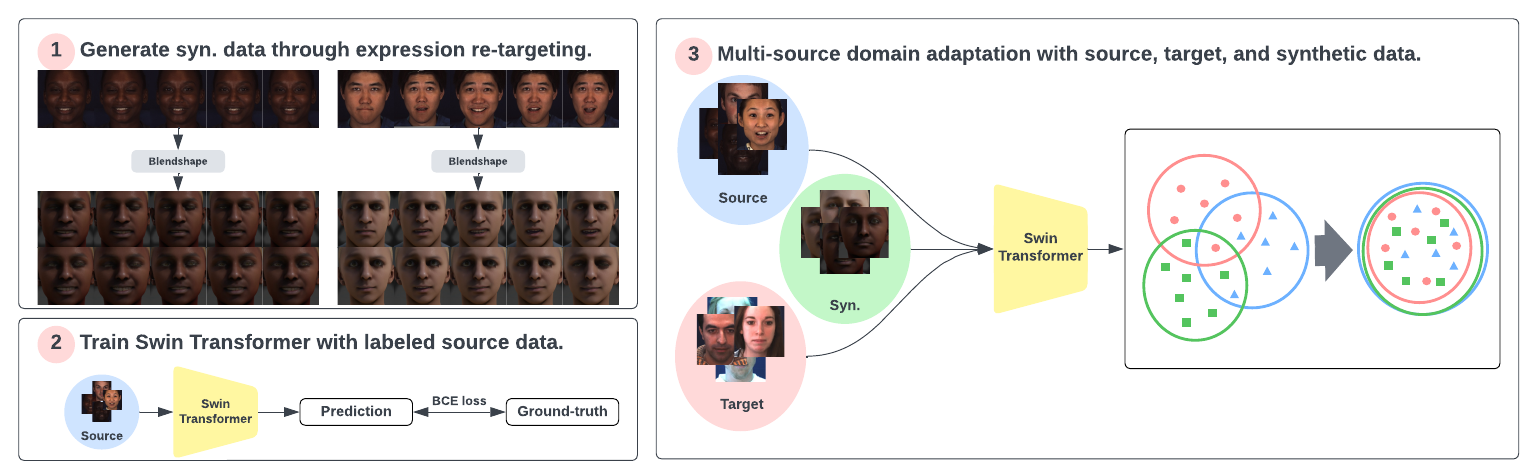}
\captionof{figure}{We leverage synthetically generated data for generalizable and fair facial action unit detection. The figure illustrates three steps: (1) We generate a diverse and balanced synthetic (syn.) dataset through expression re-targeting by transferring the expressions from real faces to synthetic avatars. (2) We train Swin Transformer with labeled source data. (3) We fine-tune Swin Transformer by aligning the feature distributions of the source, target, and our synthetic data.}
\label{fig:teaser}
}

\makeatletter
\apptocmd{\@maketitle}{\centering\insertfig}{}{}
\makeatother

\begin{document}

\title{Leveraging Synthetic Data for Generalizable and Fair Facial Action Unit Detection}
\author{
Liupei~Lu,
Yufeng~Yin,
Yuming~Gu,
Yizhen~Wu,
Pratusha~Prasad,
Yajie~Zhao,
and~Mohammad~Soleymani,
\IEEEcompsocitemizethanks{\IEEEcompsocthanksitem All authors are with the Institute for Creative Technologies, University of Southern California, Playa Vista, CA, 90094, USA \protect\\
\IEEEcompsocthanksitem The work was done when Lu was with USC Institute for Creative Technologies in 2021.}}

\markboth{}
{Leveraging Synthetic Data for Generalizable and Fair Facial Action Unit Detection}

\IEEEtitleabstractindextext{
\begin{abstract}
Facial action unit (AU) detection is a fundamental block for objective facial expression analysis. Supervised learning approaches require a large amount of manual labeling which is costly. The limited labeled data are also not diverse in terms of gender which can affect model fairness. In this paper, we propose to use synthetically generated data and multi-source domain adaptation (MSDA) to address the problems of the scarcity of labeled data and the diversity of subjects. Specifically, we propose to generate a diverse dataset through synthetic facial expression re-targeting by transferring the expressions from real faces to synthetic avatars. Then, we use MSDA to transfer the AU detection knowledge from a real dataset and the synthetic dataset to a target dataset. \edit{Instead of aligning the overall distributions of different domains, we propose Paired Moment Matching (PM2) to align the features of the paired real and synthetic data with the same facial expression. To further improve gender fairness, PM2 matches the features of the real data with a female and a male synthetic image.} Our results indicate that synthetic data and the proposed model improve both AU detection performance and fairness across genders, demonstrating its potential to solve AU detection in-the-wild.
\end{abstract}

}

\maketitle

\IEEEpeerreviewmaketitle

\IEEEraisesectionheading{\section{Introduction}}
\IEEEPARstart{F}{acial} expressions play an important role in interpersonal social communication by conveying human emotions and intentions \cite{rychlowska2017functional}. The Facial Action Coding System (FACS) is a taxonomy of facial activities that can describe expressions through anatomical action units, \eg, inner brow raiser (AU1), lip corner puller (AU 12) \cite{ekman1977facial}. Unlike facial expressions of emotions, \eg, happiness and sadness, for which there is no absolute agreement among raters \cite{barrett2019,yoder_word_2016}, FACS provides an objective measure tied to the exact movements of facial muscles to describe specific facial behaviors \cite{martinez2017automatic}.

Most of the existing work on AU detection performs within-domain evaluation, with training and testing data from the same dataset \cite{shao2018deep, shao2020jaa, ertugrul2019pattnet} while cross-domain evaluation, with training and testing data from different datasets, has not been fully investigated. Such supervised learning approaches require a large amount of manually labeled data by trained coders \cite{bartlett1999measuring}. In addition to the cost of data labeling, current AU detection datasets are not representatives of the general population. Three of the most commonly used datasets, BP4D \cite{zhang2014bp4d}, DISFA \cite{mavadati2013disfa}, and GFT \cite{girard2017sayette} suffer from imbalanced gender distributions. In this paper, we use "gender" to specify the apparent gender identity, as judged by observers and not the real gender of the subjects. The gender bias in the data can be detrimental to the fairness of the resulting machine learning models. To alleviate the problem of data scarcity and lack of diversity, synthetic AU datasets have been used in recent facial expression recognition and AU detection studies \cite{abbasnejad2017using, zhu2018emotion, niinuma2019unmasking}.

Domain adaptation (DA) aims to minimize the impact of domain shift between the source and target domains. Multi-source domain adaptation (MSDA) is a powerful extension that can transfer knowledge from different source domains to a target domain \cite{zhao2020multi}. MSDA can fully utilize data from both real and synthetic domains to learn a generalizable AU detection model.

In this paper, we propose to use synthetically generated data and multi-source domain adaptation to address the problems of the scarcity of labeled data and lack of diversity in the subjects (see Figure \ref{fig:teaser}). We propose a pipeline to generate a diverse and balanced dataset through synthetic facial expression re-targeting. Specifically, we extract the blendshape parameters from a real dataset, \ie, BP4D,  to re-target the expressions from the real videos to the synthetic avatars. The synthetic dataset consists of \textbf{60} avatars and \textbf{214,146 frames} in total. We then use multi-source domain adaptation techniques to transfer the AU detection knowledge from a real dataset and a synthetic dataset to a target dataset. 
\edit{
Existing methods such as Moment Matching for Multi-Source Domain Adaptation ($\rm M^{3}SDA$) \cite{peng2019moment} align the feature distributions of multiple domains. These methods can reduce the domain gap to some extent, however, there are three problems: (i) It is difficult to completely align the feature distributions of different domains because of their different characteristics. (ii) Matching the overall distributions does not consider any class-specific boundaries which results in reduced AU detection performance. (iii) More importantly, matching the overall distributions does not align different protected groups explicitly which may result in inferior fairness. To address these problems, we propose Paired Moment Matching (PM2) to align the features of the paired real and synthetic data with the same facial display. Instead of matching the overall feature distributions of different domains, the proposed model aims to align the data with the same class label and match the features of the real data across protected groups \ie, female and male synthetic avatars. Therefore, PM2 can keep the well-learned AU features during the domain alignment and obtain a fair representation, \ie, across genders.
}

\edit{
Extensive experiments are conducted and the results indicate that the synthetically generated data and the proposed PM2 model increase both AU detection performance and fairness across corpora. Specifically, PM2 outperforms all the single-source and multi-source domain adaptation baselines in two directions of domain adaption (BP4D to DISFA and BP4D to GFT). In addition, PM2 results in increased fairness according to the equal opportunity (EO) and statistical parity difference (SPD) between different genders compared with the baselines.
}

The major contributions of this work are as follows:
\textbf{(i)} We propose a pipeline to generate a diverse and balanced dataset for fairer machine learning models through synthetic expression re-targeting. \textbf{We release the synthetic dataset} (upon acceptance) and hope that this paper motivates others to consider fairness evaluation and further innovate to address this issue.
\textbf{(ii)} \edit{We propose Paired Moment Matching (PM2) which aligns the features of the paired real and synthetic images with the same facial expression. The proposed model aligns the data with the same class label and matches the features of the real data with both female and male synthetic avatars. Therefore, PM2 can maintain discriminative features for AU detection and obtain a fair representation in terms of gender.}
\textbf{(iii)} The experimental results indicate that our generated data and proposed model improve both cross-domain AU detection performance and fairness, demonstrating its potential to solve AU detection in a real life scenario.

\section{Related Work}
\label{sec:related}

In this section, we introduce the background and the previous work of domain adaptation and facial action unit detection.

\subsection{Domain Adaptation (DA)}
Variations in data distributions between the training and test data often result in reduced recognition performance. Domain adaptation methods \cite{long2015learning, saito2018maximum, peng2019moment} have been proposed to alleviate such problems. In particular, deep domain adaptation methods explore learning a representation that is both effective for the main task and invariant to domains \cite{wang2018deep}.

Deep Adaptation Networks (DAN) is proposed to reduce the discrepancies across source and target distributions by embedding the representations in a reproducing kernel Hilbert space where the mean embedding of different domains can be made similar \cite{long2015learning}. 
Domain-Adversarial Neural Network (DANN) \cite{ganin14} utilizes a gradient reversal layer to ensure that the feature distributions over the two domains are made similar.
\edit{
Joint Adaptation Networks (JAN) align the joint distributions of the input features and output labels in multiple domain-specific layers based on a joint maximum mean discrepancy \cite{long2017deep}. Conditional Domain Adaptation Networks (CDAN) condition the domain adaptation on discriminative information conveyed in the classifier predictions to enable alignment of the multimodal distributions \cite{long2018conditional}. Zhang \etal \cite{zhang2019bridging} propose Margin Disparity Discrepancy (MDD), which is tailored to the distribution comparison and the min-max optimization for easier training.
}
Saito \etal \cite{saito2018maximum} introduce the Maximum Classifier Discrepancy (MCD) where two classifiers are treated as discriminators and are trained to maximize the discrepancy on target samples, while a feature extractor is trained to minimize the discrepancy on target samples.

Unlike single-source domain adaptation, multi-source domain adaptation assumes that training data from multiple sources are available \cite{peng2019moment}. Ben-David \etal \cite{ben2010theory} introduce an $\mathcal{H} \Delta \mathcal{H}$-divergence between the weighted combination of source domains and target domain. Deep Cocktail Network (DCTN) \cite{xu2018deep} proposes a $k$-way domain discriminator and category classifier for digit classification and real-world object recognition. Hoffman \etal \cite{hoffman2018algorithms} present new normalized solutions with strong theoretical guarantees for the cross-entropy loss and provide a full solution for the multi-source domain adaptation problem with practical benefits. Peng \etal \cite{peng2019moment} propose Moment Matching for Multi-Source Domain Adaptation ($\rm M^{3}SDA$), which directly matches all the distributions by matching their statistical moments.

\edit{Previous methods for MSDA align the overall feature distributions by domain discriminator or minimizing the moment distance does not consider any class-specific boundaries which may result in inferior performance. While our method aligns the paired real and synthetic data with the same expression and thus maintains the discriminative features for AU detection.}

\subsection{Action Unit (AU) Detection}
A facial action unit (AU) is a facial behavior indicative of activation of an individual or a group of muscles, \eg, cheek raiser (AU 6). AUs were formalized by Ekman in Facial Action Coding System (FACS) \cite{ekman1977facial}.

Zhao \etal \cite{zhao2016deep} propose Deep Region and Multi-label Learning (DRML). The model is trained with region learning (RL) and multi-label learning (ML) and can identify more specific regions for different AUs than conventional patch-based methods.
Shao \etal \cite{shao2018deep, shao2020jaa} propose a deep learning framework named J$\hat{\text{A}}$A-Net for joint AU detection and face alignment. J$\hat{\text{A}}$A-Net uses an adaptive attention learning module to refine the attention map for each AU.

The temporal dependencies across video frames are helpful for AU detection.
Yang \etal \cite{yang2019learning} propose to learn a proxy of temporal information from the input and anchor images by estimating the hypothetical optical flow.
FAb-Net \cite{Wiles2018SelfsupervisedLO} uses facial movements from videos as the supervisory signal to learn a representation for recognizing facial attributes, through self-supervised learning.
Lu \etal \cite{lu2020self} propose a self-supervised model through learning temporal consistencies in videos. They use a triplet-based ranking approach that ranks the frames based on temporal distance from an anchor frame.

\edit{
Recent work on AU detection use graph neural networks \cite{zhang2020region, song2021hybrid, luo2022learning}.
Zhang \etal \cite{zhang2020region} utilize a heatmap regression-based approach for AU detection. The ground-truth heatmaps are defined based on the region of interest (ROI) and graph convolution is used for feature refinement.
Song \etal \cite{song2021hybrid} propose a hybrid message-passing neural network with performance-driven structures (HMP-PS). They propose a performance-driven Monte Carlo Markov Chain sampling method to generate the graph structures. Besides, hybrid message passing is proposed to combine different types of messages.
Luo \etal \cite{luo2022learning} propose an AU relationship modeling approach that learns a unique graph to explicitly describe the relationship between each pair of AUs of the target facial display.
}

Previous studies on AU detection achieve promising within-domain performance. However, the generalization ability, \ie, cross-domain performance, for AU detection has not been widely investigated.
\edit{
In this work, we refer to \textbf{within-domain} AU detection where the AU detector is trained and tested on the same datasets; and \textbf{cross-domain} AU detection where the model is trained and tested on different datasets. Specifically, the training set is the \textbf{source domain} while the test set is the \textbf{target domain}.
}

\subsubsection{Cross-domain AU Detection}
Chu \etal \cite{chu2013selective} are among the first to study cross-domain AU detection. They find that individual differences dramatically influence the classifier generalized to the unseen subjects and propose Selective Transfer Machine (STM) to personalize a general classifier in an unsupervised manner.
Ghosh \etal \cite{ghosh2015multi} propose a multi-label CNN method that learns a shared representation between multiple AUs. The experiments show that the method generalizes well to the other datasets through multi-task learning.
Later, Ertugrul \etal \cite{ertugrul2019cross, ertugrul2020crossing} demonstrate that the deep-learning-based AU detectors achieve poor cross-domain performance due to the variations in the cameras, environments, and subjects.
Tu \etal \cite{tu2019idennet} propose Identity-Aware Facial Action Unit Detection (IdenNet). IdenNet is jointly trained by AU detection and face clustering datasets that contain numerous subjects to improve the model's generalization ability.
Yin \etal \cite{yin2021self} propose to use domain adaptation and self-supervised patch localization to improve the cross-corpora performance for AU detection.

Previous studies either utilize unlabeled target data \cite{chu2013selective, yin2021self} or multi-task learning with large-scale data \cite{tu2019idennet} to improve cross-domain AU detection. In this work, we propose to generate and utilize a diverse and balanced synthetic dataset to improve the generalization ability.

\subsubsection{Fairness in Face Understanding}
\edit{
The bias in deep-learning-based models can be traced back to the imbalanced data distributions in the training set, in terms of gender, race, and age. Previous methods mitigate the bias through either data augmentation \cite{iosifidis2018dealing} or model training \cite{wang2020towards, hernandez2021deepfn, churamani2022domain}.

Iosifidis and Ntoutsi \cite{iosifidis2018dealing} generate instances from the under-represented group by over-sampling and SMOTE \cite{fernandez2018smote}.
Wang \etal \cite{wang2020towards} propose a domain-independent classifier for bias mitigation.
Hernandez \etal \cite{hernandez2021deepfn} conduct an in-depth analysis of performance differences across subjects, genders, skin types, and databases. They propose deep face normalization (DeepFN) that transfers the facial expressions of different people onto a common facial template.
Churamani \etal \cite{churamani2022domain} propose to use Domain-Incremental Learning (Domain-IL) to enhance the fairness of facial expression recognition and AU detection.

Our work combines two types of aforementioned methods. We synthetically generate additional data for data augmentation and explicitly align the different protected groups during model training.
}

\subsubsection{Synthetic Data in Facial Expression Analysis}
Synthetic data is widely used in modern computer vision including segmentation, object detection, and simulation tasks \cite{nikolenko2019synthetic, zhao2019multi, peng2015learning, tobin2017domain}. In general, many machine learning applications suffer from the data insufficiency problem, either due to small datasets or costly manual labeling \cite{nikolenko2019synthetic}. AU detection in particular is susceptible to this problem since coding AUs is highly labor-intensive. It takes over a hundred hours of training to become a certified FACS coder \cite{bartlett1999measuring}. Thus, some studies utilize synthetic facial data in AU detection and face expression analysis \cite{abbasnejad2017using, zhu2018emotion, niinuma2019unmasking}.

Zhu \etal \cite{zhu2018emotion} propose a data augmentation method using CycleGAN to find better boundaries between different unbalanced classes for expression classification. Abbasnejad \etal \cite{abbasnejad2017using} train a deep convolutional network on a generated synthetic dataset and directly fine-tune the network on the real dataset for facial expressions analysis. Recently, Wood \etal \cite{wood2021fake} propose to render training images with unprecedented realism and diversity by combining a procedurally-generated parametric 3D face model with a comprehensive library of hand-crafted assets. They demonstrate that it is possible to use synthetic data alone to perform face-related computer vision in-the-wild. Niinuma \etal \cite{niinuma2019unmasking} train a GAN-based constructor to synthesize novel images with AU labeling to solve the imbalance in AU labels and generates a large balanced dataset. \edit{The major difference between these previous studies and ours is as follows. (i) We use facial geometry re-targeting instead of GAN for synthetic data generation to avoid unstable training and mode collapse problems. (ii) We take advantage of the expression re-targeting to formulate paired real-syn data for model training.}

\begin{figure}[t]
    \centering
    \includegraphics[trim=10 10 10 10, clip, width=0.49\textwidth]{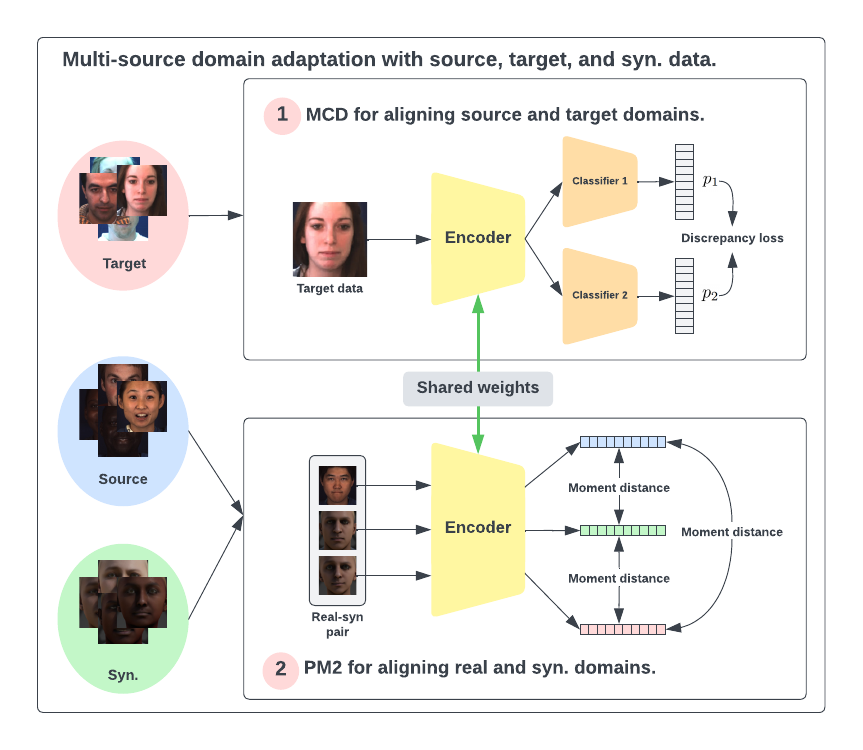}
    \caption{Overview of the proposed Paired Moment Matching (PM2). In this work, the encoder is a Swin Transformer and the classifier is an MLP. PM2 has two modules: (1) Maximum Classifier Discrepancy (MCD) aligns the source and target by minimizing the discrepancy between two classifiers on target samples. (2) PM2 aligns the real and synthetic domains by minimizing the moment distances between the real-syn pairs. Each pair contains one real image and two synthetic images generated with a male and a female avatar, sharing the same facial expression.}
    \label{fig:model}
\end{figure}

\section{Method}
In this section, we first formulate the problem and notations regarding the facial action unit detection in Section \ref{sec:problem} and provide an overview of our proposed method in Section \ref{sec:overview}. Then, we describe the model in Section \ref{sec:model}. At last, we explain the training procedure in Section \ref{sec:training} and the inference in Section \ref{sec:inference}.

\edit{
\subsection{Problem Formulation}
\label{sec:problem}
\textbf{Facial Action Unit Detection.} Given a video set $S$, for each frame $x \in S$, the goal is to detect the occurrence for each AU $y_i$ $(i=1, 2, ..., n)$ using function $\text{F}(\cdot)$.

\begin{equation}
    y_1, y_2, ..., y_n = \text{F}(x),
\end{equation}
\noindent where $n$ is the number of AUs to be detected. $y_i = 1$ if the AU is active otherwise $y_i = 0$.
}

\edit{
\subsection{Overview}
\label{sec:overview}
We propose \textbf{Paired Moment Matching (PM2)} for generalizable and fair facial action unit detection. We show the model overview in Figure \ref{fig:model}. The model comprises of an encoder $\text{E}$ (Swin Transformer \cite{liu2021swin}) and two classifiers $\text{C}_1, \text{C}_2$ (multilayer perceptron, MLP). Two classifiers are leveraged to align the distributions of the source and target domains \cite{saito2018maximum} (see Section \ref{sec:mcd}). In the meantime, we apply moment matching to align the representations of the paired real and synthetic data to further improve the model's generalization ability and fairness (see Section \ref{sec:pm2}).
}

\subsection{Model}
\label{sec:model}
\subsubsection{Base Model}
We choose Swin Transformer base \cite{liu2021swin} pre-trained on ImageNet-1K \cite{deng2009imagenet} as our base model. Swin Transformer is a hierarchical Transformer in which the representation is computed with shifted windows and can be served as a general-purpose backbone for computer vision \cite{liu2021swin}.

The base model outputs the logits for the AU occurrences through multi-task learning. The AU loss is calculated by the sum of the Binary Cross-Entropy (BCE) for every AU, as follows
\begin{equation}
\label{eq:au_loss}
    \mathcal{L}_\text{AU} = \sum_{i=1}^{n} \mathcal{L}_\text{BCE}(y_{i}, \hat{y}_{i})
\end{equation}

\noindent where $n$ is the number of AU labels to be detected, $y_i$ and $\hat{y}_{i}$ are the ground-truth and the prediction for the $i$-th AU.

\edit{
\subsubsection{Maximum Classifier Discrepancy (MCD)}
\label{sec:mcd}
To align the source and target domains, MCD aims to detect the target samples far from the source domain and then push them closer to the source. These samples are likely to be classified wrongly by the classifier trained on the source data since they are far from the source \cite{saito2018maximum}. Moreover, Saito \etal \cite{saito2018maximum} argue that these target samples outside the source are likely to be classified differently by the two distinct source classifiers. In practice, the disagreement between the two classifiers, \textbf{discrepancy}, is measured by the absolute difference of their output logits. The goal of MCD is to obtain an encoder that minimizes the discrepancy in the target samples.
\begin{equation}
\label{eq:dis_loss}
    \mathcal{L}_\text{Dis} = \sum_{x \in \mathcal{D}_T}|p_1(x)-p_2(x)|
\end{equation}

\noindent where $\mathcal{D}_T$ is the target domain, $p_i$ is the output logits of the classifier $\text{C}_i$ ($i=1,2$).
}

\begin{figure}[t]
    \centering
    \includegraphics[trim=10 10 10 10, clip, width=0.49\textwidth]{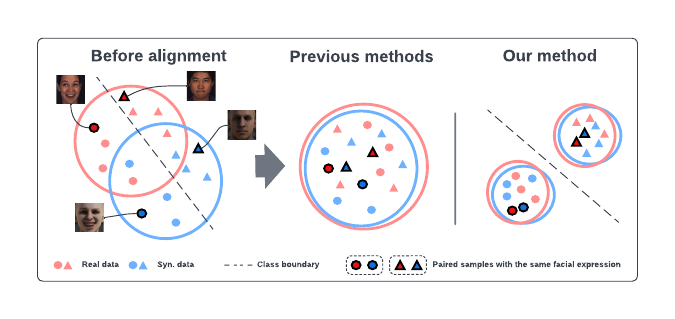}
    \caption{(Best viewed in color) Comparison of previous and proposed methods for domain alignment. \textbf{Middle:} Previous methods align the overall distributions and corrupt the class boundary. \textbf{Right:} Our method aligns the paired data with the same facial expression and maintains the class boundary.}
    \label{fig:failure_case_2}
\end{figure}

\edit{
\subsubsection{Paired Moment Matching (PM2)}
\label{sec:pm2}
To align the features of the source $\mathcal{D}_{S}$ and target $\mathcal{D}_T$, Peng \etal \cite{peng2019moment} propose $\rm M^{3}SDA$ to align the moments of their feature distributions. The objective is to minimize the moment distance
\begin{align}
    d(\mathcal{D}_S, \mathcal{D}_T) = \sum_{k=1}^{2} ||\mathbb{E}_{x \sim \mathcal{D}_S}(\text{E}(x)^k) - \mathbb{E}_{x \sim \mathcal{D}_T}(\text{E}(x)^k)||_{2}
\end{align}

\noindent where $k$ represents the $k$-th order moment of the feature (to the power of $k$), $\text{E}(x)$ is the features of image $x$ extracted from the encoder $\text{E}$.

Though $\rm M^{3}SDA$ aims to match the distributions of different domains and can reduce the domain gap to some extent, it has three problems. (i) Completely matching the features of different domains is difficult due to their different characteristics. In our case, the domain gap between the real and synthetic data is huge due to the lighting condition, head pose, and texture. (ii) Matching the overall distributions does not consider any class-specific boundaries. For example, the model may align the features of real and synthetic faces with different AUs thus resulting in inferior AU detection performance (see Figure \ref{fig:failure_case_2}). (iii) More importantly, matching the overall distributions does not align different protected groups explicitly which may result in inferior fairness (see Figure \ref{fig:failure_case_3}). In this work, we focus on fairness across genders.

To address the above problems, instead of completely aligning the overall distributions of different domains, we propose to align the paired real and synthetic faces with the same facial expression. Our synthetic data is generated through facial expression re-targeting, thus, we can easily obtain the paired data. In this way, the model will avoid aligning the images with different AU labels which may corrupt the well-learned AU representations. To further improve gender fairness, for each real face, we generate two synthetic images with the same facial expression, one with a female avatar and another one with a male avatar. By aligning the real face to these two synthetic images, the model can obtain a fair representation in terms of gender.

\begin{figure}[t]
    \centering
    \includegraphics[trim=10 10 10 10, clip, width=0.49\textwidth]{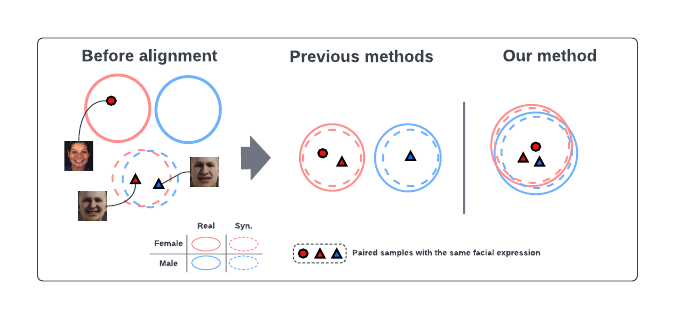}
    \caption{(Best viewed in color) Comparison of previous and proposed methods for bias mitigation. \textbf{Left:} The real domain has gender bias thus female and male distributions are not aligned. While the synthetic (syn.) data is balanced and the gender distributions are matched. \textbf{Middle:} Though previous methods match the real and synthetic domains well, they fail to align the gender distributions. \textbf{Right:} Our method aligns each real image with a male and a female avatar. The method explicitly aligns the real and synthetic, female and male domains.}
    \label{fig:failure_case_3}
\end{figure}

Formally, given an input image $x$, through facial expression re-targeting, we generate two synthetic images with the same facial expression, $x_f$ (with a female avatar) and $x_m$ (with a male avatar). We minimize the moment distance between the paired real and synthetic data to align the domains and thus improve the model generalization ability and fairness.
\begin{align}
    d(x_1, x_2) = \sum_{k=1}^{2} ||(\text{E}(x_1)^k) - (\text{E}(x_2)^k)||_{2}
\end{align}
\begin{align}
\label{eq:pm2_loss}
    \mathcal{L}_\text{PM2} = \frac{1}{3}\sum_{x \in \mathcal{D}_S} (d(x, x_f)+d(x, x_m)+d(x_f, x_m))
\end{align}
}

\subsection{Training}
\label{sec:training}
The overall training procedure for the proposed PM2 includes three steps which are performed alternately until convergence.

\textbf{Step 1} We train the encoder $\text{E}$ and the classifiers $\text{C}_1, \text{C}_2$ to classify the real source samples correctly and align the real and synthetic domains in the meantime.
\begin{equation}
    \mathcal{L}_1 = \frac{1}{2}(\mathcal{L}_{\text{AU}_1}+\mathcal{L}_{\text{AU}_2})+\beta \cdot \mathcal{L}_\text{PM2}
\end{equation}

\noindent where $\mathcal{L}_{\text{AU}_i}$ is the AU loss (Equation \ref{eq:au_loss}) for the classifier $\text{C}_i$ on the real source domain ($i=1, 2$), $\mathcal{L}_{\text{PM2}}$ is the moment distance loss (Equation \ref{eq:pm2_loss}) between the paired real and synthetic data, and $\beta$ is a hyper-parameter.

\textbf{Step 2} The encoder $\text{E}$ is fixed and the classifiers $\text{C}_1, \text{C}_2$ are trained to maximize the discrepancy on the target samples.
\begin{equation}
    \mathcal{L}_2 = \frac{1}{2}(\mathcal{L}_{\text{AU}_1}+\mathcal{L}_{\text{AU}_2})+\beta \cdot \mathcal{L}_\text{PM2} - \alpha \cdot \mathcal{L}_\text{Dis}
\end{equation}

\noindent where $\mathcal{L}_{\text{Dis}}$ is the discrepancy loss (Equation \ref{eq:dis_loss}) on the target domain and $\alpha$ is a hyper-parameter.

\textbf{Step 3} The classifiers $\text{C}_1, \text{C}_2$ are fixed, and the encoder $\text{E}$ is trained to minimize the discrepancy on the target domain.
\begin{equation}
    \mathcal{L}_3 = \alpha \cdot \mathcal{L}_\text{Dis}
\end{equation}

\begin{figure*}[t]
    \centering
    \includegraphics[trim=130 0 130 0, clip, width=\textwidth]{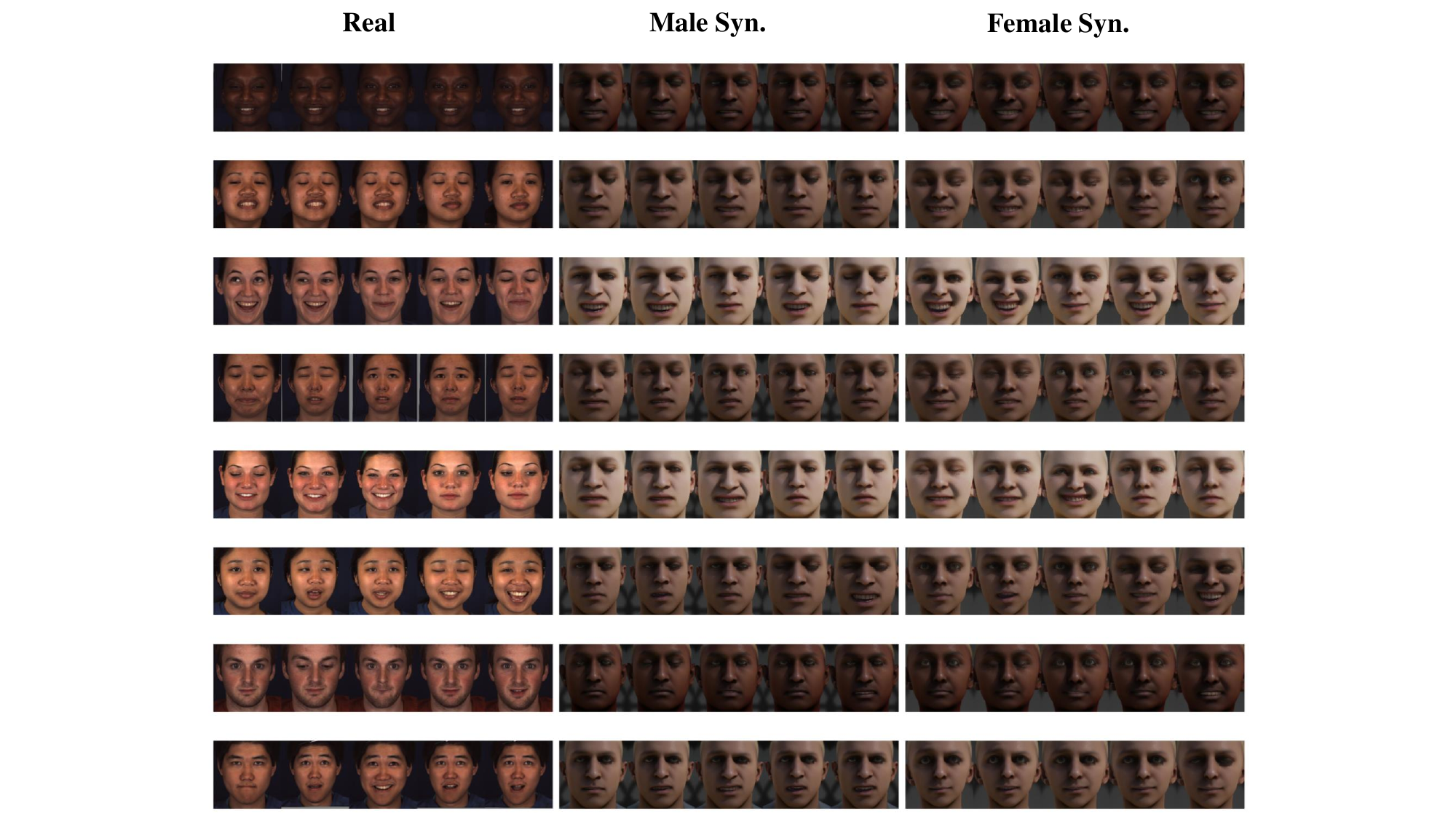}
    \caption{Examples of synthetically generated data through expression re-targeting. Real images come from BP4D. For each real subject, we generate the synthetic data with two avatars, one male and one female. The synthetic data has similar facial expressions as the real images.}
    \label{fig:syn_data}
\end{figure*}

\subsection{Inference}
\label{sec:inference}
During the testing phase, we calculate the average output logits of the two classifiers as the final prediction.
\begin{equation}
    p(x) = \frac{1}{2}(p_1(x)+p_2(x))
\end{equation}

\noindent where $p_i$ is the output logits of the classifier $\text{C}_i$ ($i=1,2$).

\section{Experiments}
\begin{figure*}[t]
\centering
    \begin{subfigure}{0.245\textwidth}
    \centering
    \includegraphics[trim=255 140 255 140, clip, width=\textwidth]{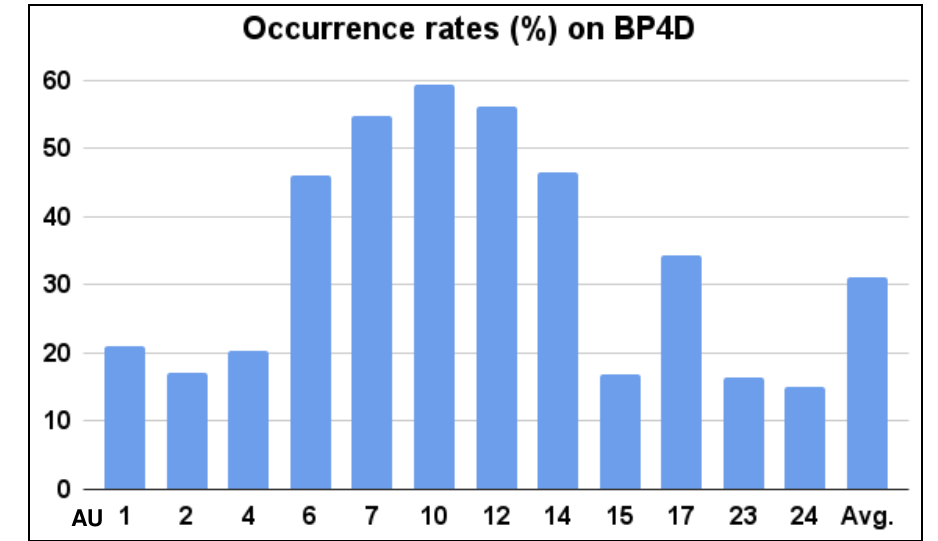}
    \caption{BP4D.}
    \label{fig:bp4d}
    \end{subfigure}
    \begin{subfigure}{0.245\textwidth}
    \centering
    \includegraphics[trim=255 140 255 140, clip, width=\textwidth]{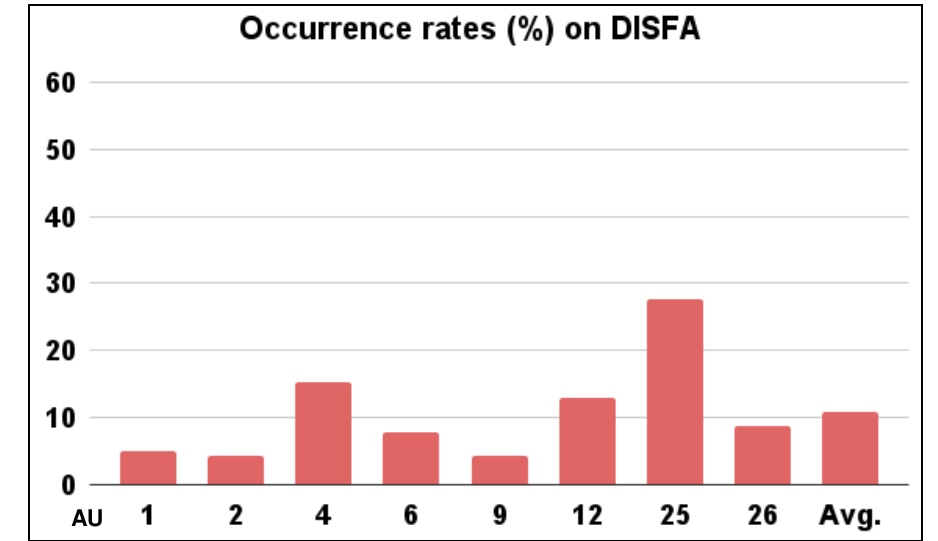}
    \caption{DISFA.}
    \label{fig:disfa}
    \end{subfigure}
    \begin{subfigure}{0.245\textwidth}
    \centering
    \includegraphics[trim=255 140 255 140, clip, width=\textwidth]{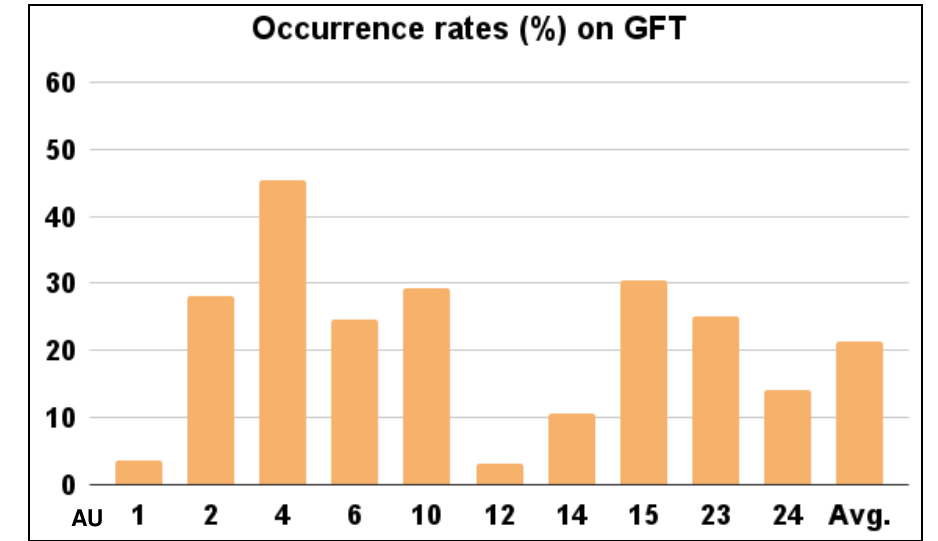}
    \caption{GFT.}
    \label{fig:gft}
    \end{subfigure}
    \begin{subfigure}{0.245\textwidth}
    \centering
    \includegraphics[trim=255 140 255 140, clip, width=\textwidth]{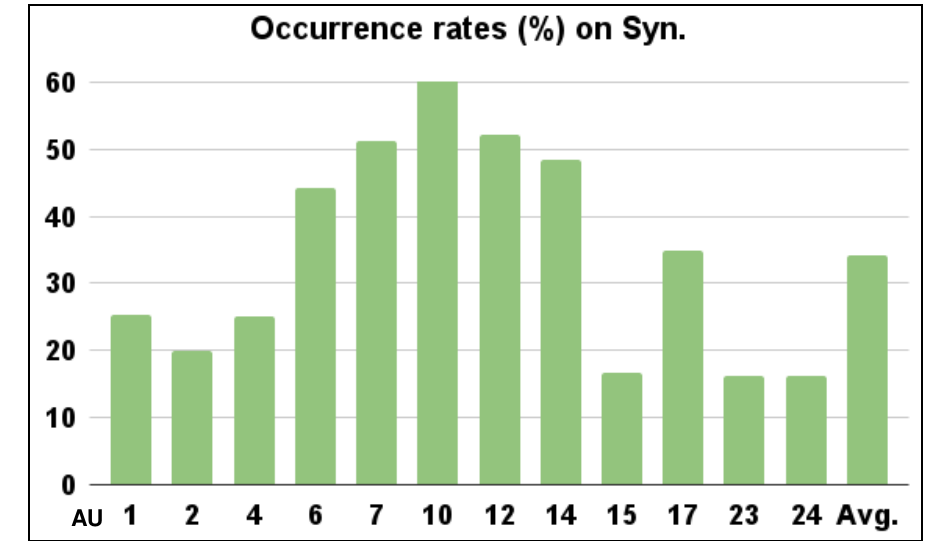}
    \caption{Syn.}
    \label{fig:syn}
    \end{subfigure}
    \caption{Occurrence rates of AU labels for different datasets. BP4D has the highest occurrence rate on average while DISFA and GFT are severely imbalanced.}
    \label{fig:base_rate}
\end{figure*}

\begin{figure}[t]
    \centering
    \includegraphics[trim=40 0 40 0, clip, width=0.3\textwidth]{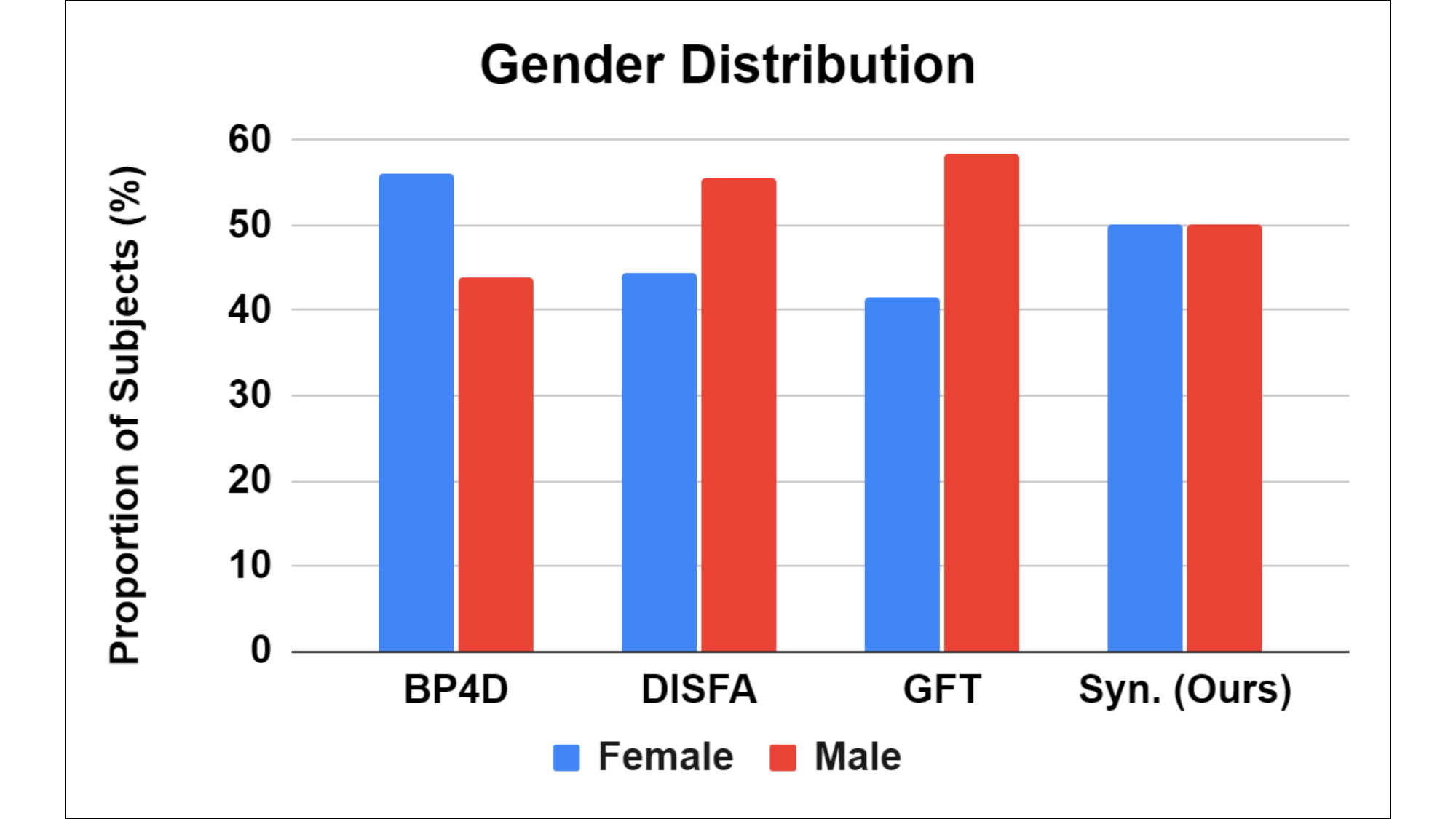}
    \caption{Gender distributions for different datasets. All the real datasets suffer from imbalanced gender distributions while our synthetically generated data (Syn.) is \textbf{completely balanced}.}
    \label{fig:gender}
\end{figure}

In this section, we conduct extensive experiments to show the effectiveness of our proposed method compared to the existing work. We first introduce the datasets in Section \ref{sec:data}. The implementation and training details are given in Section \ref{sec:implementation}. Finally, in Section \ref{sec:results}, we show the experimental results including within- and cross-domain evaluations, fairness evaluation, ablation study, and case study.

\subsection{Data}
\label{sec:data}
In this work, we have two source domains, namely real and synthetic data, and one target domain for the multi-source domain adaptation. The synthetic data is generated by facial geometry re-targeting to increase the subject diversity and fairness.

\subsubsection{Real Data}
Three widely used benchmarks for AU detection are selected, namely, Binghamton–Pittsburgh 4D Spontaneous Expression Database (BP4D) \cite{zhang2014bp4d}, Denver Intensity of Spontaneous Facial Actions (DISFA) \cite{mavadati2013disfa}, and Sayette Group Formation Task Spontaneous Facial Expression Database (GFT) \cite{girard2017sayette}.

\textbf{BP4D} consists of videos from 41 subjects with around 146,000 frames. Each frame comes with AU occurrence labels and 49 landmarks detected by a CLM method \cite{saragih2011deformable}. Each frame has labels for 12 AUs (1, 2, 4, 6, 7, 10, 12, 14, 15, 17, 23, and 24).

\textbf{DISFA} includes videos from 27 subjects, with a total of 130,000 frames. Each frame is annotated with AU intensities from 0 to 5 and 66 landmarks detected by an AAM \cite{cootes1998active}. Each frame has labels for eight AU intensities (1, 2, 4, 6, 9, 12, 25, and 26). Following \cite{zhao2016deep, li2019self,shao2018deep}, we map AU intensity greater than 1 to the positive class.

\textbf{GFT} contains 96 subjects from 32 three-person groups with unscripted social interaction. Each subject has about 1,800 annotated frames. Video frames are annotated for ten AUs (1, 2, 4, 6, 10, 12, 14, 15, 23, and 24) and 49 landmarks detected by ZFace \cite{Jeni15FG_ZFace}.

BP4D is recorded under strictly controlled lighting conditions and thus an ``easier'' dataset compared to DISFA and GFT, with a higher average reported performance \cite{shao2018deep, shao2019facial, shao2020jaa}. Thus, we choose BP4D as the source domain and DISFA and GFT as the target domain.

\subsubsection{Synthetic Data}
\label{sec:syn_data}
With the facial AU configurations from the real data, \ie, BP4D, we can generate multiple synthetic avatars with the expression transferred to the avatars. Specifically, we first detect the facial landmarks and extract the blendshape parameters by the Faceware Studio\footnote{\url{https://facewaretech.com/software/studio/}}. Then, we use the parameters to map the tracked expressions to the synthetic avatars to generate the same AU configurations. The face attributes and the generated textures are based on Li \etal \cite{li2020learning}. The rendering is performed using the Unreal Engine\footnote{\url{https://www.unrealengine.com/}}. We utilize a customized 3D morphable model (3DMM) to have a controllable subject appearance. The 3D morphable model has two adjustable components: Control shape and Shader Parameter. Control shape decomposes 3D face shape attributes like gender, cheek, and jaw by principal component analysis (PCA), and Shader Parameter allows us to synthesize various skin colors through different shading on texture. In the end, we use the Unreal Engine to render the morphed avatars to generate 2D images with the corresponding facial expressions.

The extracted blendshape is the linear weighted sum of facial muscle actions, which ensures the facial AU is kept. Moreover, such transfer is aligned at the frame level. Thus, the correctness of AU transfer is very likely via such transformations, and we annotate the synthetic data with the corresponding real labels.

\begin{table}[t]
\centering
\caption{Implementation details. Values of the hyper-parameters for within- and cross-domain evaluations.}
\begin{tabular}{l|c|c}
\toprule
\rowcolor{Gray}
Hyper-parameter & Within-domain & Cross-domain \\
\midrule
\# epochs & 30 & 10 \\
learning rate & 1e-4 & 1e-6\\
batch size & 128 & 32 \\
\midrule
$\alpha$ for $\mathcal{L}_\text{Dis}$ & - & 0.3 \\
$\beta$ for $\mathcal{L}_\text{PM2}$ & - & 0.5 \\
\bottomrule
\end{tabular}
\label{tab:hyperparamter}
\end{table}

\edit{
We randomly select 30 subjects from the real dataset, \ie, BP4D. For each real subject, we perform expression re-targeting for all their frames with two synthetic avatars, one male and one female.} Overall, the synthetic dataset consists of \textbf{60 avatars} with a total of \textbf{214,146 frames}.

Figure \ref{fig:syn_data} provides more visual examples for expression re-targeting from the real data to the synthetic avatars. Specifically, the synthetic data can transfer the eyelid movement well, but the eyebrow changes are less obvious. This is due to the limitation of our 3D morphable model capacity. As for the AUs in the lower face, the lips movements (raiser/puller) of synthetic images are also consistent with the source for obvious muscle movements. Overall, the synthetically generated data has similar facial expressions as the real images.

\begin{table*}[t]
\centering
\caption{Within-domain evaluation for the base model (Swin Transformer \cite{liu2021swin}) on different datasets. 
\textbf{F1-score (\% $\uparrow$)} is the evaluation metric. GFT has a lower average F1-score than BP4D and DISFA.}
\begin{tabular}{l|ccccccccccccccc|c}
\toprule
\rowcolor{Gray}
AU & 1 & 2 & 4 & 6 & 7 & 9 & 10 & 12 & 14 & 15 & 17 & 23 & 24 & 25 & 26 & \textbf{Avg.} \\
\midrule
BP4D \cite{zhang2014bp4d} & 50.6 & 50.1 & 63.2 & 77.9 & 80.3 & - & 83.1 & 88.3 & 57.9 & 49.6 & 61.4 & 44.5 & 44.2 & - & - & 62.6 \\
\midrule
DISFA \cite{mavadati2013disfa} & 58.6 & 56.5 & 72.5 & 47.6 & - & 51.3 & - & 76.4 & - & - & - & - & - & 95.2 & 65.5 & 65.5 \\
\midrule
GFT \cite{girard2017sayette} & 41.9 & 57.5 & 3.9 & 73.1 & - & - & 65.0 & 78.8 & 24.1 & 46.0 & - & 44.1  & 53.6 & - & - & 48.8 \\
\bottomrule
\end{tabular}
\label{tab:within_domain}
\end{table*}

\begin{table*}[t]
\centering
\footnotesize
\caption{Cross-domain evaluation with different models in different directions of domain adaptation. Swin stands for Swin Transformer \cite{liu2021swin}. \textcolor{cyan}{$*^{\text{src}}$} means the model is trained with the source data and directly transferred to the target domain. \textcolor{red}{$*^{\text{tgt}}$} means the model is both trained and tested with the target data. \textbf{F1-score (\% $\uparrow$)} is the evaluation metric. The best results are shown in \textcolor{orange}{\textbf{orange}}. The proposed PM2 has the highest average F1-score, indicating the best generalization ability.}
\scalebox{0.94}{\begin{tabular}{l|ccccc|a|cccccccccc|a|a}
\toprule
\rowcolor{Gray}
Direction & \multicolumn{6}{c|}{BP4D $\rightarrow$ DISFA} & \multicolumn{11}{c|}{BP4D $\rightarrow$ GFT} & \textbf{Avg.} \\
\midrule
\rowcolor{Gray}
AU & 1 & 2 & 4 & 6 & 12 & \textbf{Avg.} & 1 & 2 & 4 & 6 & 10 & 12 & 14 & 15 & 23 & 24 & \textbf{Avg.} & \\
\midrule
DRML\textcolor{cyan}{$^{\text{src}}$} \cite{zhao2016deep} & 9.4 & 4.2 & 23.7 & 18.2 & 27.2 & 16.5 & 7.5 & 18.3 & 5.6 & 55.1 & 48.8 & 56.2 & 5.9 & 1.8 & 14.6 & 14.6 & 22.8 & 19.7 \\
J$\hat{\text{A}}$A-Net\textcolor{cyan}{$^{\text{src}}$} \cite{shao2020jaa} & 11.0 & 15.1 & 25.2 & 19.9 & 38.7 & 22.0 & 10.4 & 19.9 & 6.6 & 54.4 & 47.6 & 63.5 & 5.9 & 8.6 & 18.0 & 17.7 & 26.9 & 24.4 \\
ResNet\textcolor{cyan}{$^{\text{src}}$} \cite{he2016deep} & 16.7 & 13.7 & 30.4 & 20.5 & 27.5 & 21.8 & 8.5 & 3.4 & 10.2 & 48.1 & 36.3 & 45.6 & 6.5 & 7.9 & 0.1 & 23.4 & 19.0 & 20.4 \\
Swin\textcolor{cyan}{$^{\text{src}}$} \cite{liu2021swin} & 21.6 & 18.8 & 46.8 & 32.9 & 50.7 & 34.2 & 15.9 & 35.5 & 12.4 & 54.2 & 52.0 & 68.9 & 7.9 & 2.5 & 13.9 & 26.8 & 29.0 & 31.6 \\
\midrule
DAN \cite{long2015learning} & 26.0 & 23.3 & 47.5 & 33.8 & 46.0 & 35.3 & 19.6 & 39.6 & 8.7 & 59.5 & 62.8 & 69.2 & 10.1 & 26.6 & 35.8 & 38.1 & 37.0 & 36.2 \\
DANN \cite{ganin14} & 27.4 & 28.9 & 42.2 & 39.8 & 52.8 & 38.2 & 24.6 & 28.8 & 7.9 & 67.6 & 68.5 & 74.2 & 11.5 & 25.8 & 32.4 & 38.3 & 38.0 & 38.1 \\
JAN \cite{long2017deep} & 26.7 & 26.1 & 51.9 & 32.1 & 42.7 & 35.9 & 27.3 & 41.3 & 8.6 & 65.7 & 64.5 & 69.7 & 11.9 & 23.1 & 34.7 & 40.4 & 38.7 & 37.3 \\
CDAN \cite{long2018conditional} & 28.0 & 23.0 & 39.7 & 37.0 & 51.2 & 35.8 & 30.1 & 25.6 & 8.2 & 68.9 & 68.8 & 75.1 & 11.5 & 25.3 & 30.1 & 39.6 & 38.3 & 37.1 \\
MCD \cite{saito2018maximum} & 31.4 & 28.7 & 44.8 & 39.1 & 51.3 & 39.1 & 28.4 & 34.3 & 10.1 & 67.0 & 70.3 & 75.2 & 12.8 & 14.0 & 16.2 & 40.8 & 36.9 & 38.0 \\
$\rm M^{3}SDA$ \cite{peng2019moment} & 43.3 & 26.3 & 49.8 & 40.9 & 50.5 & 42.1 & 27.8 & 38.4 & 8.8 & 68.9 & 68.7 & 73.9 & 11.8 & 17.9 & 30.3 & 40.7 & 38.7 & 40.4 \\
\midrule
\textbf{PM2 (Ours)} & 46.0 & 27.6 & 49.8 & 35.6 & 54.6 & \textcolor{orange}{\textbf{42.7}} & 29.0 & 32.0 & 9.0 & 67.9 & 70.2 & 74.2 & 11.4 & 18.4 & 37.1 & 44.1 & \textcolor{orange}{\textbf{39.3}} & \textcolor{orange}{\textbf{41.0}} \\
\midrule
Swin\textcolor{red}{$^{\text{tgt}}$} \cite{liu2021swin} & 58.6 & 56.5 & 72.5 & 47.6 & 76.4 & 62.3 & 41.9 & 57.5 & 3.9 & 73.1 & 65.0 & 78.8 & 24.1 & 46.0 & 44.1 & 53.6 & 48.8 & 55.6 \\
\bottomrule
\end{tabular}}
\label{tab:cross_domain}
\end{table*}

\subsubsection{Data Observation}
To statistically evaluate the fairness and diversity of subjects for the aforementioned datasets, we report the distributions of AU labels and genders.

\textbf{Label Distribution.} The occurrence rates of AU labels for different datasets are given in Figure \ref{fig:base_rate}. We observe that many of the AUs occur infrequently. Specifically, BP4D has the highest occurrence rate on average. DISFA and GFT are severely imbalanced as some AUs occur in less than 15\% of frames (AU 1, 2, 6, 9, 12, 26 in DISFA, and AU 1, 12, 14, 24 in GFT). The AU occurrence rate for the synthetic data is similar to BP4D as it originates from the dataset.

\textbf{Gender Distribution.} Figure \ref{fig:gender} shows the gender distributions. All three real datasets are imbalanced while our synthetic data is generated to have an equal number of apparent female/male subjects.

\subsection{Implementation Details}
\label{sec:implementation}
We use a machine with two Intel(R) Xeon(R) Gold 5218 CPUs @ 2.30GHz with eight NVIDIA Quadro RTX8000 GPUs for all the experiments.

All methods are implemented in Python 3.9.13 and PyTorch \cite{paszke17} 1.10.1. \edit{The related packages and versions are: torchvision 0.11.2, numpy 1.23.1, scikit-learn 1.1.2, scipy 1.9.1, pillow 9.2.0, pandas 1.4.4, and opencv-python 4.6.0.66.} Code and model weights are available, for the sake of reproducibility\footnote{htttp://-url-to-appear-upon-acceptance}.

\edit{All frames are cropped and aligned using the facial landmarks detected by dlib \cite{dlib09}. Each image is first resized into $256 \times 256$. We apply random horizontal flip and random cropping during training for data augmentation.}

All models are trained with an AdamW optimizer \cite{loshchilov2017decoupled}. Table \ref{tab:hyperparamter} provides the values per (hyper-)parameter. Thresholds for AU detection for all AUs are set to 0.5, for all experiments. This means that if the output logit is greater than $0.5$, then the prediction is active otherwise inactive.

\subsection{Experimental Results}
\label{sec:results}
We perform both within- and cross-domain experiments. Our main focus is to improve AU detection performance across corpora indicative of the generalization ability of the model and its fairness (across genders).
\begin{table*}[t]
\centering
\caption{Fairness evaluation between different gender groups according to the \textbf{equal opportunity (EO \% $\uparrow$)}. Swin stands for Swin Transformer \cite{liu2021swin}. The best results are shown in \textcolor{orange}{\textbf{orange}}. The proposed PM2 has the highest equal opportunity, indicating the best fairness.}
\scalebox{0.94}{\begin{tabular}{l|ccccc|a|cccccccccc|a|a}
\toprule
\rowcolor{Gray}
Direction & \multicolumn{6}{c|}{BP4D $\rightarrow$ DISFA} & \multicolumn{11}{c|}{BP4D $\rightarrow$ GFT} & \textbf{Avg.} \\
\midrule
\rowcolor{Gray}
AU & 1 & 2 & 4 & 6 & 12 & \textbf{Avg.} & 1 & 2 & 4 & 6 & 10 & 12 & 14 & 15 & 23 & 24 & \textbf{Avg.} & \\
\midrule
Swin\textcolor{cyan}{$^{\text{src}}$} \cite{liu2021swin} & 69.3 & 32.3 & 74.5 & 97.2 & 89.4 & 72.6 & 80.3 & 96.0 & 48.9 & 99.2 & 72.1 & 88.3 & 48.1 & 26.6 & 90.8 & 92.3 & 74.2 & 73.4 \\
\midrule
DAN \cite{long2015learning} & 62.6 & 48.3 & 90.9 & 74.6 & 95.1 & 74.3 & 64.7 & 99.1 & 56.4 & 88.7 & 84.5 & 88.8 & 65.9 & 46.3 & 87.6 & 97.9 & 78.0 & 76.2 \\
DANN \cite{ganin14} & 65.9 & 44.3 & 88.2 & 72.4 & 98.4 & 73.8 & 64.1 & 97.8 & 98.8 & 85.7 & 89.8 & 94.8 & 63.6 & 52.6 & 63.3 & 93.1 & 80.4 & 77.1 \\
JAN \cite{long2017deep} & 58.4 & 39.3 & 99.8 & 77.6 & 84.4 & 71.9 & 77.6 & 91.4 & 46.3 & 82.3 & 82.0 & 84.1 & 61.4 & 45.4 & 79.5 & 92.7 & 74.3 &  73.1 \\
CDAN \cite{long2018conditional} & 66.7 & 50.7 & 81.4 & 76.3 & 99.3 & 74.9 & 74.4 & 90.8 & 85.3 & 84.6 & 90.6 & 94.2 & 65.0 & 53.9 & 67.1 & 94.5 & 80.0 & 77.5 \\
MCD \cite{saito2018maximum} & 73.9 & 45.3 & 93.0 & 72.1 & 96.7 & 76.2 & 58.5 & 87.4 & 72.1 & 90.2 & 92.2 & 96.3 & 49.0 & 77.5 & 77.6 & 91.2 & 79.2 & 77.7 \\
$\rm M^{3}SDA$ \cite{peng2019moment} & 98.3 & 39.1 & 70.7 & 76.9 & 97.2 & 76.4 & 67.2 & 85.9 & 41.5 & 86.9 & 90.1 & 95.3 & 55.2 & 52.0 & 97.9 & 90.1 & 76.2 & 76.3 \\
\midrule
\textbf{PM2 (Ours)} & 61.0 & 78.7 & 83.9 & 73.7 & 99.9 & \textcolor{orange}{\textbf{79.5}} & 77.1 & 71.3 & 87.9 & 86.1 & 94.2 & 93.3 & 55.2 & 52.4 & 96.3 & 93.6 & \textcolor{orange}{\textbf{80.7}} & \textcolor{orange}{\textbf{80.1}} \\
\bottomrule
\end{tabular}}
\label{tab:fair}
\end{table*}

\begin{table*}[t]
\centering
\caption{Fairness evaluation between different gender groups in terms of the \textbf{statistical parity difference (SPD \% $\downarrow$)}. Swin stands for Swin Transformer \cite{liu2021swin}. The best results are shown in \textcolor{orange}{\textbf{orange}}. The proposed PM2 has the lowest SPD, indicating the best fairness.}
\scalebox{1}{\begin{tabular}{l|ccccc|a|cccccccccc|a|a}
\toprule
\rowcolor{Gray}
Direction & \multicolumn{6}{c|}{BP4D $\rightarrow$ DISFA} & \multicolumn{11}{c|}{BP4D $\rightarrow$ GFT} & \textbf{Avg.} \\
\midrule
\rowcolor{Gray}
AU & 1 & 2 & 4 & 6 & 12 & \textbf{Avg.} & 1 & 2 & 4 & 6 & 10 & 12 & 14 & 15 & 23 & 24 & \textbf{Avg.} & \\
\midrule
Swin\textcolor{cyan}{$^{\text{src}}$} \cite{liu2021swin} & 7.5 & 18.5 & 8.0 & 22.1 & 1.0 & 11.4 & 8.7 & 9.1 & 11.5 & 1.8 & 4.7 & 13.1 & 0.2 & 0.2 & 2.8 & 7.0 & 5.9 & 8.7 \\
\midrule
DAN \cite{long2015learning} & 1.1 & 3.7 & 5.0 & 9.3 & 10.1 & 5.8 & 8.1 & 7.2 & 3.1 & 2.5 & 1.2 & 0.8 & 2.4 & 0.9 & 3.1 & 4.9 & 3.4 & 4.6 \\
DANN \cite{ganin14} & 0.6 & 0.4 & 4.7 & 4.6 & 6.3 & 3.3 & 4.9 & 3.8 & 5.5 & 0.4 & 2.1 & 4.3 & 0.3 & 0.6 & 2.4 & 5.0 & 2.9 & 3.1 \\
JAN \cite{long2017deep} & 1.2 & 6.5 & 5.5 & 13.0 & 14.7 & 8.2 & 4.0 & 5.1 & 3.7 & 0.1 & 0.0 & 2.1 & 0.1 & 0.9 & 0.7 & 3.0 & 2.0 & 5.1 \\
CDAN \cite{long2018conditional} & 0.5 & 0.2 & 5.4 & 8.4 & 6.4 & 4.2 & 4.1 & 2.9 & 5.1 & 1.1 & 1.8 & 5.1 & 1.0 & 0.0 & 0.4 & 5.1 & 2.7 & 3.5 \\
MCD \cite{saito2018maximum} & 3.0 & 0.2 & 4.9 & 8.4 & 6.6 & 4.6 & 2.9 & 3.1 & 6.4 & 0.2 & 3.3 & 5.4 & 3.8 & 1.2 & 1.9 & 3.1 & 3.1 & 3.9 \\
$\rm M^{3}SDA$ \cite{peng2019moment} & 7.0 & 2.7 & 5.2 & 3.1 & 9.4 & 5.5 & 4.6 & 2.9 & 5.6 & 3.1 & 3.9 & 3.7 & 0.4 & 1.3 & 1.3 & 2.4 & 2.9 & 4.2 \\
\midrule
\textbf{PM2 (Ours)} & 1.0 & 0.2 & 1.7 & 2.8 & 3.3 & \textcolor{orange}{\textbf{1.8}} & 0.7 & 0.6 & 5.0 & 1.0 & 0.4 & 1.6 & 0.2 & 0.1 & 2.1 & 1.7 & \textcolor{orange}{\textbf{1.3}} & \textcolor{orange}{\textbf{1.6}} \\
\bottomrule
\end{tabular}}
\label{tab:spd}
\end{table*}

\subsubsection{Within-domain Evaluation}
In terms of BP4D, DISFA, and the synthetic dataset, the base model is evaluated using subject-independent three-fold cross-validation, with two folds for training and one fold for validation, similar to \cite{he2016deep, shao2020jaa}. For GFT, following the original training/validation/test partitions \cite{girard2017sayette}, we use 60, 18, and 18 subjects for training, validation, and test respectively. F1-score is the evaluation metric.

Table \ref{tab:within_domain} shows the within-domain results for the base model on the three real datasets. From the table, we observe that GFT has a lower average F1-score than BP4D and DISFA. We suspect it is because BP4D and DISFA have better video quality than GFT.

\subsubsection{Cross-domain Evaluation}
We perform two directions of unsupervised domain adaptation. We set BP4D as the real source domain and DISFA/GFT as the target domain. Specifically, we employ all the labeled source data and all the unlabeled target data for training, and all the target data for testing. For the multi-source domain adaptation baseline and our own method, we also train the models with the synthetic data. When DISFA is the target domain, we evaluate the five common AU labels (1, 2, 4, 6, 12), and we evaluate the ten common AUs (1, 2, 4, 6, 10, 12, 14, 15, 23, 24) when GFT is the target domain. F1-score ($\uparrow$) is the evaluation metric. The results of the cross-domain evaluations are shown in Table \ref{tab:cross_domain}.

\edit{We implement four base models, \ie, DRML \cite{zhao2016deep}, J$\hat{\text{A}}$A-Net \cite{shao2020jaa}, ResNet-18 \cite{he2016deep}, and Swin Transformer base \cite{liu2021swin} and report their direct transfer performance. In Table \ref{tab:cross_domain}, we observe that Swin Transformer has the highest cross-domain performance in both directions of domain adaptation. We suspect it is due to the following reasons. (i) Compared with DRML and J$\hat{\text{A}}$A-Net which are trained from scratch, Swin Transformer is first pre-trained on ImageNet-1K \cite{deng2009imagenet} and thus has better generalization ability. (ii) Compared with ResNet, Swin Transformer has more training parameters (121M vs 11M), thus Swin Transformer has a higher capacity.

We also report the supervised performance of the Swin Transformer on the target domain. Compared to the performance of direct transfer, we observe a large performance drop in both directions of domain adaptation (\textbf{24.0\%} F1-score decrease on average). This demonstrates the challenging nature of cross-domain AU detection and the importance of developing generalizable AU detection.}

For the rest of the experiments, we employ Swin Transformer as the base model for domain adaptation. We implement five single-source domain adaptation methods and one multi-source domain adaptation model as the baselines, \ie, DAN \cite{long2015learning}, DANN \cite{ganin14}, JAN \cite{long2017deep}, CDAN \cite{long2018conditional}, MCD \cite{saito2018maximum}, and $\rm M^{3}SDA$ \cite{peng2019moment}.

We observe that domain adaptation can mitigate the performance drop to some extent. All the domain adaptation baselines improve the cross-domain performance in both directions. In particular, MCD \cite{saito2018maximum} and DANN \cite{ganin14} are the best two single-source domain adaption models while DAN \cite{long2015learning} achieves the worst performance in terms of the average F1-score. \edit{$\rm M^{3}SDA$ \cite{peng2019moment} and our own method are built upon MCD but utilize the synthetic data for multi-source domain adaptation. Compared with MCD, these two models can further improve the cross-domain performance in both directions, demonstrating the effectiveness of our synthetically generated data. The proposed PM2 achieves the highest average F1-score, indicating the best generalization ability. The major improvement of PM2 over other baselines comes from AU 12 (lip corner puller) from BP4D to DISFA and AU 23 (lip tightener), AU 24 (lip pressor) from BP4D to GFT. We suspect it is because our synthetic data are more consistent with the original source in terms of lip movements than the movements in the upper face. Future work should introduce more rigs around eyebrows to better mimic the animation. Though the improvement of PM2 over $\rm M^{3}SDA$ is not significant, PM2 has better fairness across genders (see the next section).}
\begin{table*}[t]
\centering
\caption{Ablation study for PM2 with different loss weights. \textbf{F1-score (\% $\uparrow$)} is the evaluation metric. The best results are shown in \textcolor{orange}{\textbf{orange}}. DT stands for direct transfer. PM2 achieves the best performance with joint training of $\mathcal{L}_\text{Dis}$ and $\mathcal{L}_\text{PM2}$.}
\scalebox{0.91}{\begin{tabular}{l|cc|ccccc|a|cccccccccc|a|a}
\toprule
\rowcolor{Gray}
& \multicolumn{2}{c|}{Direction} & \multicolumn{6}{c|}{BP4D $\rightarrow$ DISFA} & \multicolumn{11}{c|}{BP4D $\rightarrow$ GFT} & \textbf{Avg.} \\
\midrule
\rowcolor{Gray}
Method & $\alpha$ & $\beta$ & 1 & 2 & 4 & 6 & 12 & \textbf{Avg.} & 1 & 2 & 4 & 6 & 10 & 12 & 14 & 15 & 23 & 24 & \textbf{Avg.} & \\
\midrule
DT & 0 & 0 & 21.6 & 18.8 & 46.8 & 32.9 & 50.7 & 34.2 & 15.9 & 35.5 & 12.4 & 54.2 & 52.0 & 68.9 & 7.9 & 2.5 & 13.9 & 26.8 & 29.0 & 31.6 \\
\midrule
\multirow{2}{*}{$\mathcal{L}_\text{Dis}$ only} & 1 & 0 & 31.4 & 28.7 & 44.8 & 39.1 & 51.3 & 39.1 & 28.7 & 30.3 & 10.6 & 63.2 & 69.4 & 72.2 & 12.6 & 9.4 & 11.5 & 30.1 & 33.8 & 36.5 \\
& 0.3 & 0 & 31.6 & 25.6 & 45.3 & 35.0 & 49.0 & 37.3 & 28.4 & 34.3 & 10.1 & 67.0 & 70.3 & 75.2 & 12.8 & 14.0 & 16.2 & 40.8 & 36.9 & 37.1 \\
\midrule
\multirow{2}{*}{$\mathcal{L}_\text{PM2}$ only} & 0 & 1 & 31.6 & 29.7 & 42.0 & 31.7 & 50.5 & 37.1 & 21.4 & 10.6 & 7.9 & 67.9 & 70.9 & 75.3 & 10.4 & 16.7 & 30.0 & 39.8 & 35.1 & 36.1 \\
& 0 & 0.5 & 32.8 & 28.1 & 47.1 & 35.0 & 53.1 & 39.2 & 22.3 & 13.0 & 8.0 & 67.1 & 70.3 & 74.6 & 10.3 & 17.0 & 29.2 & 38.5 & 35.0 & 37.1 \\
\midrule
\multirow{4}{*}{PM2} & 1 & 1 & 29.6 & 23.2 & 50.0 & 42.7 & 57.8 & 40.7 & 26.1 & 15.0 & 8.3 & 67.4 & 69.6 & 74.7 & 10.8 & 12.9 & 26.0 & 37.9 & 34.8 & 37.8 \\
& 1 & 0.5 & 27.5 & 29.4 & 45.3 & 38.8 & 58.4 & 39.9 & 25.4 & 23.4 & 9.3 & 61.2 & 65.4 & 71.0 & 11.5 & 16.1 & 31.8 & 37.1 & 35.2 & 37.6 \\
& 0.3 & 1 & 37.1 & 29.2 & 46.9 & 37.3 & 55.1 & 41.1 & 26.0 & 18.7 & 8.1 & 68.1 & 70.5 & 75.5 & 10.4 & 19.2 & 32.7 & 40.9 & 37.0 & 39.1 \\
& 0.3 & 0.5 & 46.0 & 27.6 & 49.8 & 35.6 & 54.6 & \textcolor{orange}{\textbf{42.7}} & 29.0 & 32.0 & 9.0 & 67.9 & 70.2 & 74.2 & 11.4 & 18.4 & 37.1 & 44.1 & \textcolor{orange}{\textbf{39.3}} & \textcolor{orange}{\textbf{41.0}} \\
\bottomrule
\end{tabular}}
\label{tab:ablation}
\end{table*}

\begin{figure*}[t]
\centering
\footnotesize
\begin{tabular}{m{2.3cm} m{2.8cm} m{2.8cm} m{0.6cm} m{2.8cm} m{2.8cm} m{0.6cm}}
\toprule
Image & \includegraphics[width=.15\textwidth]{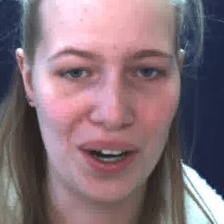} & \includegraphics[width=.15\textwidth]{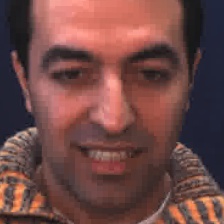} & & \includegraphics[width=.15\textwidth]{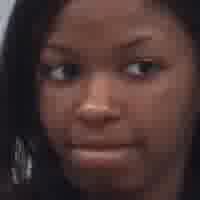} & \includegraphics[width=.15\textwidth]{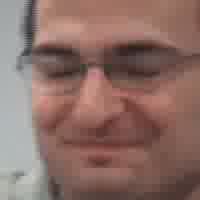} & \\
\end{tabular}
\begin{tabular}{L{2.3cm} C{2.8cm} C{2.8cm} C{0.6cm} C{2.8cm} C{2.8cm} C{0.6cm}}
\midrule
Active AU & Inner Brow Raiser & Inner Brow Raiser & & Lip Pressor & Lip Pressor & \\
Gender & Female & Male & & Female & Male & \\
\midrule
Output logit & & & Diff. & & & Diff. \\
$\rm M^{3}SDA$ \cite{peng2019moment} & \textcolor{red}{0.49 ($\tikzxmark$)} & \textcolor{red}{0.44 ($\tikzxmark$)} & 0.05 & \textcolor{ForestGreen}{0.59 ($\tikzcmark$)} & \textcolor{red}{0.47 ($\tikzxmark$)} & 0.12 \\
\textbf{PM2 (Ours)} &  \textcolor{ForestGreen}{\textbf{0.59 ($\tikzcmark$)}} & \textcolor{ForestGreen}{\textbf{0.60 ($\tikzcmark$)}} & \textcolor{ForestGreen}{\textbf{0.01}} & \textcolor{ForestGreen}{\textbf{0.73 ($\tikzcmark$)}} & \textcolor{ForestGreen}{\textbf{0.75 ($\tikzcmark$)}} & \textcolor{ForestGreen}{\textbf{0.02}} \\
\bottomrule
\end{tabular}
\caption{Case analysis on $\rm M^{3}SDA$ \cite{peng2019moment} and our method. The output logits for the AU occurrences are reported. Logit greater than $0.5$ means the predicted AU is active. Lower logit difference (Diff.) indicate better fairness. Our method achieves better AU detection performance and gender fairness.}
\label{fig:case_study}
\end{figure*}

\subsubsection{Fairness Evaluation}
Two metrics are applied to evaluate the fairness of different models, \ie, \textbf{equal opportunity (EO $\uparrow$)} \cite{hardt2016equality, churamani2022domain} and \textbf{statistical parity difference (SPD $\downarrow$)} \cite{dwork2012fairness} across genders (female and male). We report and compare the performances of the base model and all the domain adaptation baselines with our own method. The results are shown in Table \ref{tab:fair} for EO and Table \ref{tab:spd} for SPD.

\textbf{Equal opportunity} suggests that given a sensitive attribute, the classifier's performance should be equal for all groups of that attribute \cite{hardt2016equality}. In practice, Churamani \etal \cite{churamani2022domain} calculate the metric through the ratio of the lowest performance for a sensitive attribute with respect to the highest performance for that sensitive attribute.
\begin{equation}
    \text{EO}_i = \frac{\min(\text{f}(\hat{y}_i, y_i|a=0), \text{f}(\hat{y}_i, y_i|a=1))}{\max(\text{f}(\hat{y}_i, y_i|a=0), \text{f}(\hat{y}_i, y_i|a=1))}
\end{equation}

\noindent where $\text{f}(\hat{y}_i, y_i|a)$ is the F1-score for the $i$-th AU with the sensitive attribute $a$, $a=1$ means the subject belongs to the minority group otherwise the majority group.

\edit{
From Table \ref{tab:fair}, we observe that the direct transfer (Swin\textcolor{cyan}{$^{\text{src}}$}) achieves the lowest EO, indicating the worst fairness. Compared with direct transfer, the domain adaptation baselines can improve fairness except for JAN. MCD obtains the highest average EO.
}

\textbf{Statistical parity} recognizes a predictor to be unbiased if the prediction $\hat{y}$ is independent of the protected attribute $a$ so that $\text{P}(\hat{y}|a) = \text{P}(\hat{y})$ \cite{dwork2012fairness}. Here, the same proportion of each population is classified as positive. Deviations from statistical parity are measured by the statistical parity difference (SPD)
\begin{equation}
    \text{SPD}_i=|\text{P}(\hat{y}_i=1|a=1)-\text{P}(\hat{y}_i=1|a=0)|
\end{equation}

\noindent where $\text{P}(\hat{y}_i=1|a)$ means the predicted occurrence rate for the $i$-th AU with the sensitive attribute $a$, $a=1$ means the subject belongs to the minority group otherwise the majority group.

\edit{
Table \ref{tab:spd} shows the SPD results. Similar to EO, we observe that the direct transfer (Swin\textcolor{cyan}{$^{\text{src}}$}) achieves the highest SPD, indicating the worst fairness while the domain adaptation baselines outperform the direct transfer.

Compared with MCD, $\rm M^{3}SDA$ utilizes the balanced synthetic data for training. $\rm M^{3}SDA$ improves the AU detection performance, however, it becomes more biased in terms of gender fairness for both EO and SPD. We speculate that the model matches the overall distributions and does not explicitly align the female and male domains.

Compared with $\rm M^{3}SDA$, our method aims to match the features of the real face with both female and male avatars. Such explicit domain alignment helps PM2 to obtain a fair representation in terms of gender. For both EO and SPD evaluations, the proposed PM2 achieves superior fairness than the other baselines, demonstrating the effectiveness of training with diverse synthetic data and paired moment matching.
}

\edit{
\subsubsection{Ablation Study}
We investigate the individual contributions of different components of PM2. We report these results in Table \ref{tab:ablation}. Specifically, we evaluate PM2 with different loss weights, \ie, $\alpha$ for $\mathcal{L}_\text{Dis}$ and $\beta$ for $\mathcal{L}_\text{PM2}$. In particular, $\mathcal{L}_\text{Dis}$ helps PM2 to align the source and target domains while $\mathcal{L}_\text{PM2}$ matches the distributions of the real and synthetic data.

When both $\alpha$ and $\beta$ are $0$, PM2 is reduced to the direct transfer, achieving the lowest performance in both directions of domain adaptation. Compared with the direct transfer, adding $\mathcal{L}_\text{Dis}$ or $\mathcal{L}_\text{PM2}$ to the objective improves the overall performance. In addition, from BP4D to GFT, compared with $\mathcal{L}_\text{Dis}$ only, $\mathcal{L}_\text{PM2}$ only achieves superior performance for AU 23 (lip tightener) but inferior results for AU 1 and AU 2 (inner brow raiser and outer brow raiser). We suspect it is because the eyebrow behaviors are not well transferred from the real faces onto avatars.

Finally, combining both loss terms obtains the best performance, demonstrating the effectiveness of joint training with source, target, and synthetic data. In particular, we have consistent findings that the model achieves higher performance when $\alpha = 0.3$ and $\beta = 0.5$. This could be due to the trade-off between AU detection and domain alignment. The discriminative features for AU detection may be corrupted if the model puts too much emphasis on domain alignment.
}

\edit{
\subsubsection{Case Study}
Figure \ref{fig:case_study} shows the comparison between the output logits of $\rm M^{3}SDA$ and our model on two pairs of samples. Each pair has two images. The first pair comes from DISFA while the second one is from GFT. The first pair has active AU 1 (inner brow raiser) while the second one has active AU 24 (lip pressor). When the output logit is greater than $0.5$, the model prediction is active otherwise inactive.

We observe that $\rm M^{3}SDA$ fails in three of the cases while our method accurately predicts the AUs, indicating that our method has a strong generalization ability. Moreover, our method achieves a lower logit difference than $\rm M^{3}SDA$. The results show that our method is fairer across genders.
}

\section{Discussions}
Synthetic facial expression re-targeting allows us to generate a diverse and balanced dataset using avatars. However, the current data generation pipeline is a multi-stage process. First, the facial landmarks and blendshape are extracted by the Faceware Studio from real images using a graphical interface. Then, we apply extracted blenshape to the avatars to transfer the expression. Finally, we utilize a 3D morphabale model to change the physical identities of the avatar. Due to the lack of the software API interface, all three steps are not scripted in an end-to-end fashion and are operated manually, resulting in cases of frame drop and requiring extensive manual work. In addition, as mentioned in Section \ref{sec:syn_data}, eyebrow behaviors are not well transferred from real human images to avatars, this is due to the limitation of our 3D morphable model capacity. For future work, the entire process could be automated for larger scope usage, and more rigs should be introduced around eyebrows to better mimic the animation.

In this work, we only focus on gender fairness. We would like to generate more paired synthetic data for other attributes, \eg, race, and age in future work.

On the model level, due to the difficulty of cross-corpus facial AU detection, there is still a huge gap between the source and target domain performance. Though the proposed PM2 improves fairness, the improvement for the cross-domain evaluation is not significant. Future work should consider addressing these problems in the method design, optimization, and choice of the loss function.

\section{Conclusions}
In this paper, we leverage synthetic data to build a more generalizable facial AU detection model. We find that gender biases exist in widely used AU detection datasets resulting in reduced fairness of AU detection. To address these problems, we propose to generate a diverse and balanced dataset through synthetic facial expression re-targeting. \edit{We propose Paired Moment Matching (PM2) to align the features of the paired real and synthetic data with the same facial expression. Instead of completely matching the overall distributions of different domains, the proposed model aims to align the data with the same class label and match the features of the real face with both female and male synthetic avatars. Therefore, PM2 can keep the well-learned AU features during the domain alignment and obtain a fair representation (across genders).} Extensive experiments are conducted whose results indicate that the synthetically generated data and the proposed PM2 model increase both AU detection performance and fairness across corpora, demonstrating its potential to solve AU detection in a real-life scenario.

For our future work, we would like to generate a larger-scale synthetic dataset across different subject attributes and appearance, \eg, race and age.

\section*{Acknowledgments}
Research was sponsored by the Army Research Office and was accomplished under Cooperative Agreement Number W911NF-20-2-0053. The views and conclusions contained in this document are those of the authors and should not be interpreted as representing the official policies, either expressed or implied, of the Army Research Office or the U.S. Government. The U.S. Government is authorized to reproduce and distribute reprints for Government purposes notwithstanding any copyright notation herein.

\bibliographystyle{ieeetr}
\bibliography{ref}

\end{document}